# Automatic crack classification by exploiting statistical event descriptors for Deep Learning

Giulio Siracusano1,* Francesca Garescì2, Giovanni Finocchio3,4,* Riccardo Tomasello5, Francesco Lamonaca6, Carmelo Scuro7, Mario Carpentieri5, Massimo Chiappini4,* and Aurelio La Corte1

[1] Department of Electric, Electronic and Computer Engineering, University of Catania, Viale Andrea Doria 6, I-95125 Catania, Italy
[2] Department of Engineering, University of Messina, C. di Dio, S. Agata, I-98166 Messina, Italy
[3] Department of Mathematical and Computer Sciences, Physical Sciences and Earth Sciences, University of Messina, C. di Dio, S. Agata, I- 98166 Messina, Italy
[4] Istituto Nazionale di Geofisica e Vulcanologia (INGV), Via di Vigna Murata 605, I-00143 Roma, Italy
[5] Department of Electrical and Information Engineering, Politecnico di Bari, via E. Orabona 4, I-70125 Bari, Italy
[6] Department of Informatics, Modeling, Electronics and System Engineering, University of Calabria, Via P. Bucci I-87036. Rende (CS), Italy
[7] Department of Physics, University of Calabria, Via P. Bucci I-87036. Rende (CS), Italy
* Correspondence: giuliosiracusano@gmail.com (G.S.); gfinocchio@unime.it (G.F.) ; massimo.chiappini@ingv.it (M.Ch.)

## Abstract

In modern building infrastructures, the chance to devise adaptive and unsupervised data-driven health monitoring systems is gaining in popularity due to the large availability of big data from low-cost sensors with communication capabilities and advanced modeling tools such as Deep Learning. The main purpose of this paper is to combine deep neural networks with Bidirectional Long Short Term Memory and advanced statistical analysis involving Instantaneous Frequency and Spectral Kurtosis to develop an accurate classification tool for tensile, shear and mixed modes originated from acoustic emission events (cracks). We investigated on effective event descriptors to capture the unique characteristics from the different types of modes. Tests on experimental results confirm that this method achieves promising classification among different crack events and can impact on the design of future on structural health monitoring (SHM) technologies. This approach is effective to classify incipient damages with 92% of accuracy, which is advantageous to plan maintenance.

## 1. Introduction

The aim of structural health monitoring (SHM) is to achieve efficient and cost-effective structural reliability and sustainability through integrated health management and prognosis. The area of the SHM receiving most attention in literature [1–9] is extraction of data features allowing to distinguish between the undamaged and damaged structures.

The continue growing of Internet-of-Things (IoT) and digital twin has brought many new sensing mechanisms suitable for SHM. In particular, acoustic emission (AE) monitoring [10,11] is becoming an established method for feature extraction. It is based on simultaneous analysis of several parameters, such as vibration amplitude and frequency, with the characteristics of acoustic wave originated from a crack. AE signals are usually investigated using the Moment Tensor Analysis (MTA) [12] and/or the calculation of AE parameters [2]. AE parameters, such as average frequency (AF), rise time (RT), duration (DUR), rise angle (RA), peak amplitude as well as energy (ENE), have shown to be effective to classify both tensile and shear failures [2,8,13–17]. During solicitations, it is well known that the propagation distance of the acoustic wave and the quality of the propagation path (i.e. presence of defects, air gaps) have an impact on the calculation of AE parameters, such as AF and RA [6,18], especially in the presence of damage [4]. However, there is no such consolidated procedure for the identification of mixed-mode events [19] occurring mostly at the transition from a regime having mainly tensile crack events to the other characterized by the majority of shear crack events. A recent study [14] has pointed out that data quality can be compromised by untreated disturbances related to the measuring system (i.e. the sensor-induced distortion) or the environment (e.g. scattering attenuation due to damage accumulation, viscous damping, inhomogeneity of concrete, structural defects of internal specimen, etc…). Therefore, those disturbances should be carefully taken into account and minimized [10,20] for reliability.

Here, we present a procedure to address the above described problems in order to have an automatic and robust classification of crack events in AE data. We use different statistical operators, i.e. Instantaneous Frequency, Spectral Entropy and Spectral Kurtosis, which have already been effective for characterizing a variety of non-stationary signals [21,22] combined with deep learning (DL) neural networks. Instead of applying DL methodology to image input data of concrete surfaces [23,24], in this paper, we use DL to classify different AE solicitations represented by time varying signals.

The chance to properly characterize the temporal behavior of AE events using DL depends on the type of information that contributes to the learning process. For example, traditional AE parameters, which generally provide a single number as a result (e.g. AF, RA, ENE), are not a favorable choice for feeding a DL network, because they cause an increase in the learning time and decrease in the overall performance [25].

Temporal series classification is a common analysis problem which requires to identify a functional dependence between the set of possible time series and the finite set of classes using a training set with known classes. There are several works which illustrate that DL is suitable for classification problem and can outperform other algorithms [26,27] which encourage further work in this direction. There exist DL models, such as recurrent neural networks (RNNs) that are designed specifically for processing sequential data, and thus could be applied for time series.

Here, we have built a RNN model constituted of bidirectional long short-term memory units (Bi-LSTM) [28] to properly classify different classes of crack events. The main motivation for this choice is that Bi-LSTMs are: (i) easier to train than RNNs [29], (ii) effective in the classification and prediction of time-series [25], and (iii) capable of learning long-term dependencies [30,31]. The performance of the DL neural network is evaluated on a dataset of AE measurements and compared against other DL-based techniques in terms of classification accuracy. Furthermore, results are comparable to other achievements [32–34] when applying DL to detecting faults in rotating machinery. We wish to underline that an additional benefit of our framework is that can be embedded into SHM systems for the lifetime monitoring of large-scale structures. The paper is organized as follows. Section 2 provides information about the acquisition system and the experimental setup. Section 3 describes the proposed framework. Sections 4 and 5 present results, discussions, and conclusions, respectively.

## 2. Multi-sensors Acquisition System and experimental setup

The AE-based methodology relies on the investigation of the elastic energy as generated from a crack formation. In concrete, the standardization procedure for the characterization of AE data is active [35], aiming at proposing a well-established setup for sensors and measurement procedures. With regard to this, previous studies [36,37] have proposed a multi-triggered acquisition system (AS), which takes into account emerging standards, enabling both high sampling frequencies and reduced storage requirements. Such types of AS are attracting a growing interest and becoming widespread [10,38,39] to modern infrastructure with long service life. Specifically, for the setup used here, an asymmetric arrangement [37] of five AE transducers has been used (Fig. 1 (a) illustrates a representation of the acquisition system with a sketch of the sensors layout S1-S5). Signals from all transducers are pre-amplified by 40 dB, digitized and collected by a multi-channel system. Each channel (Ch1-Ch5) is directly coupled to a single sensor (S1-S5). The software, in cooperation with the acquisition board, automatically triggers the recording of such relevant events [40].

### *2.1. Experimental Setup*

The experimental setup consists of:

- One hydraulic press with a closed loop governing system with 5000 kN connected to the AS to control and record the load-displacement diagram;
- Piezoelectric transducers, R15$\alpha$, with a peak sensitivity of 69 V/(m/s), resonant frequency 150 kHz, and directionality ±1.5 dB [41];
- Controlling hardware appliance constituted by multiple Logic Flat Amplifier Trigger generator (L-FAT) and DAta acQuisition boards (DAQ) NI-6110 with four input channels each, 12-bit resolution, and sampling frequency $f_{AS}$ = 5 Msample/s wherein a channel (Ch) is directly associated to each transducer;

A complete description of the acquisition system can be found in Ref. [37]. The experimental tests were conducted on a set of concrete cubic specimens with dimension 15 x 15 x 15 cm$^3$ without any steel reinforcement. Such specimens were cured for 28 days with temperature 20 ± 2 °C and relative humidity equal to 95% according to the norm UNI 6132-72 [42].

The AE transducers were arranged along the vertical sides of the specimen [37]. After calibration, we have investigated 50 specimens having different compressive strengths, $R_{ck}$, ranging from 25 to 45 MPa (cylinder/cube compressive strengths). The load was applied to one face using uniaxial compression, while the other ones are maintained fixed. The constant displacement rate was 0.1 mm/min [43] until failure. Figure 1 (b) provides some of the experimental compression load curves obtained for the concrete specimens. All the corresponding load curves are characterized by an elastic regime (linear region) with the maximum at the peak load (inflection point) [44]. After crossing that point, the material deformations become irreversible (plastic regime) and the non-linear response indicates that the specimen is significantly damaged and close to the collapse [37,45]. During the compression test, the AE events were detected, collected and processed to be used for DL training and testing purposes. Measurement dataset is available online (see Appendix A).

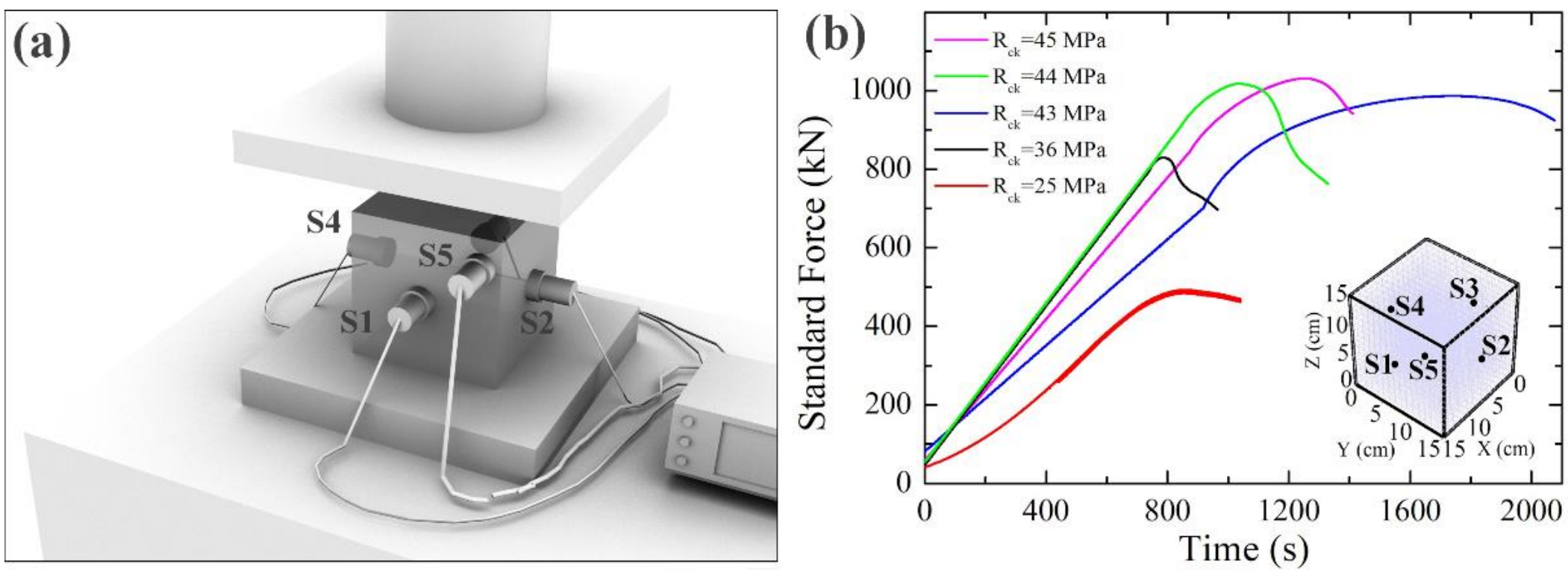

**Figure 1.** (a) A sketch of the acquisition system. (b) Experimental load-vs-time diagrams for the concrete specimens having different values of $R_{ck}$ ranging between 25 and 45 MPa. Inset: Schematics of the sensors layout.

## 3. Framework for the real-time classification of Acoustic Emission data

Figure 2 shows the schematic block diagram of the processing pipeline. It comprises a number of waveform preprocessing blocks [37] performing : (I) transducer's transfer function removal (TFR), (II) Hilbert-Huang transform (HHT) and (III.1) feature extraction. Once preprocessed, the output data feed the (III.2) DL-based [37,46] processing block used for the classification of AE signals. From a given input, each block on the pipeline performs different I/O operations and it is numbered according to sequential tasks. In the waveform preprocessing blocks, any raw transducer time-domain measurement $r_{k,q}(t)$ related to the $q$-th AE (q = 1…Q) captured from the $k$-th channel ($k$ = 1…5 in our system) is processed through blocks from (I) to (III.2). In (I), $r_{k,q}(t)$ is deconvoluted using TFR [10] with the sensing acquisition system, thus generating a reconstructed signal $s_{k,q}(t)$. Then in (II), $s_{k,q}(t)$ is processed by means of the HHT [47], in order to perform denoising and detrending, generating refined signal $\tilde{s}_{k,q}(t)$. Subsequently, we evaluate, against all the available channels, the solicitation having the highest energy E, $\hat{s}_q(t)$, such that $\hat{s}_q(t) = \max_k \left\{ E_{\tilde{s}_{k,q}} \right\}$. $\hat{s}_q(t)$ is then used to feed the next block of the pipeline.

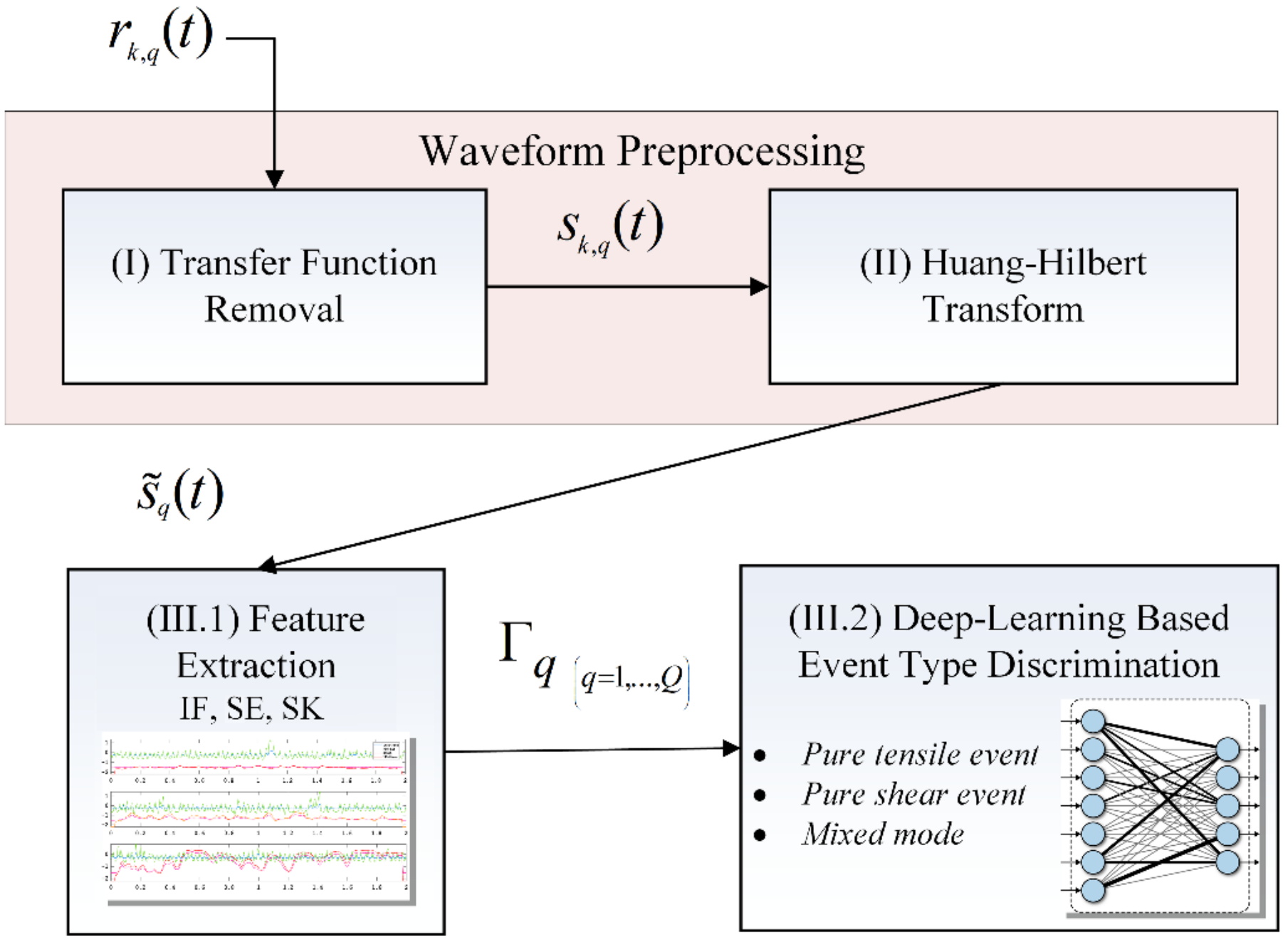

**Figure 2.** Block diagram of the framework for damage classification implemented in this work.

In block (III.1), for a given $q$-th AE, we extract different statistical properties (described in Section 3.2 and Table I), hereinafter named as event descriptors (ED): instantaneous frequency (IF) [48], spectral entropy (SE) [21], spectral kurtosis (SK) [22] that are arranged into the matrix $\Gamma_q$:

$$\Gamma_q = \left\{ IF\left(\hat{s}_q(t)\right), SE\left(\hat{s}_q(t)\right), SK\left(\hat{s}_q(t)\right) \right\} \tag{1}$$

This array of features, $\Gamma_q$, is used to populate the input dataset, $\mathbf{\Gamma}$, such that:

$$\mathbf{\Gamma} = \left[ \Gamma_q \right]_{q=1,\ldots,Q} \tag{2}$$

In $\Gamma_q$, each Event Descriptor (ED) is represented by a series composed of $N_{ED}$ elements which represent the dynamic of the chosen statistical function once it is calculated from the input data $\left(\hat{s}_q(t)\right)$. Considering that we collected Q=15000 AE events, where each event is of 2 ms duration (10.000 samples), we extracted the properties to have a 67 samples discretization ($N_{ED}$ = 67). Therefore, the available dataset $\Gamma$ has a size of Q x 3 x $N_{ED}$ = 15000 x 3 x 67.

Finally, the block (III.2) performs DL-based classification of AE events by analyzing Γ through a deeply stacked Bi-LSTM architecture (described in Sections 3.3-3.4) and providing the recognized category of event as output. Here, the input data are used to automatically discriminate among crack solicitations by solving a Multi-class classification, which is the problem of categorizing instances into precisely one of more than two classes.

### *3.1. Characterization of different crack events*

It is well known that the shape of an AE signal is intrinsically representative of the nature of the underlying fracture modes which generate tensile, shear or mixed-mode deformations. Therefore, the timely characterization of such cracking events can act as a safe measure against the final collapse [6]. As demonstrated in [37], we rely upon the versatility of HHT to analyze AE signals and to highlight the main characteristics of such different types of solicitations. In Figure 3 (a)-(c), we illustrate a comparison among such tensile, shear and mixed-mode events by focusing on their independent components and plotting the intrinsic mode functions (IMFs) extracted with the HHT from the signals data set.

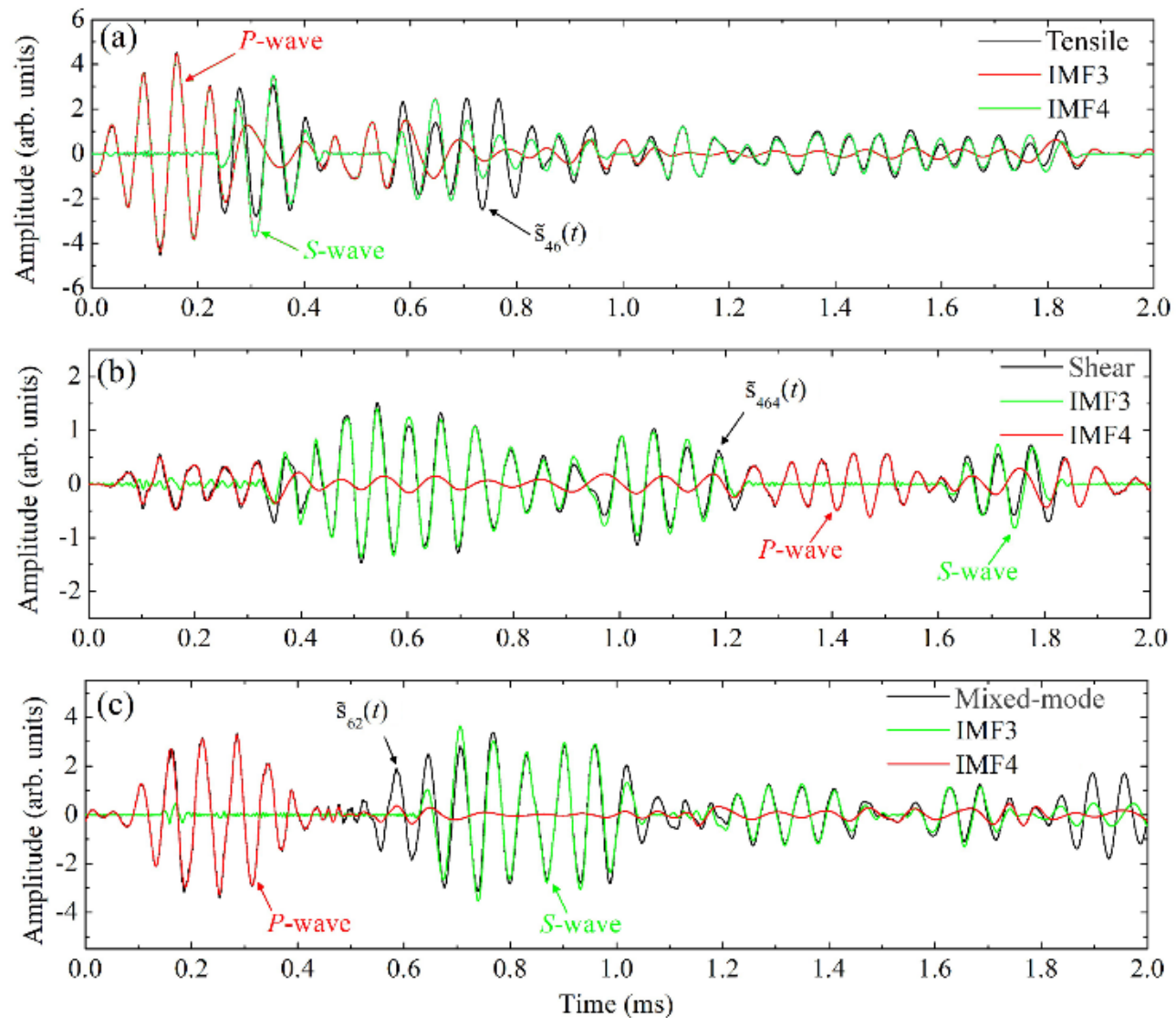

**Figure 3.** Examples of the time domain trace (black line) P- (red line) and S- (green line) waves as extracted with the HHT from AE waveforms related to different failure events: (a) tensile, (b) shear, and (c) mixed-mode, respectively.

Figure 3 (a) represents an AE time trace as emitted during a tensile event (Mode I). When a tensile event takes place, the moving sides of the crack lead to a transient volumetric change in the material. As a consequence, most of the energy is released in the form of P-waves, which are faster [49], whereas only a limited amount is transferred into S-waves, which are slower. As expected, the AE wave (black line) is constituted by a large P-wave (red line) [50], followed by a S-wave (green line) having a smaller amplitude [51]. For tensile events, the RT is short [52], and then a high RA is observed. Figure 3 (b) shows an AE time trace as emitted during a shear event (Mode II). In this case, the shape (and not the volume) of the material in proximity of the crack changes. In this case, the shape (and not the volume) of the material in proximity of the crack changes. In this case, the shear crack (black line) emits most of the energy in the form of *S*-waves (green line) and only a small amount in the form of *P*-waves (red line). In terms of AE parameters, this cracking event is described by a long RT and a short RA [18]. Figure 3 (c) displays an example of time trace of a mixed mode (Mode III) typically observed during the transition from tensile dominated regime to shear dominated one. Mixed modes are originated from a combination of tensile and shear cracks [53–55] and are important for the wide range of civil engineering [13,55–57]. Such events can be described by a more balanced combination of *P*- and *S*-waves [53]. Nonetheless, the mixed-modes cannot be properly identified by calculating AE parameters, such as RT or RA [10]. Therefore, a more flexible approach is required to discriminate among different classes of crack events. The proposed classification of the main fracture modes is based on the approach as described in Ref. [37] and confirmed in [58–61]. Our experimental measurements show AE data characterized by the generation of tensile and shear events as well as mixed-mode.

### *3.2. Analysis of Acoustic Emission Events using Feature Extraction*

In order to envisage both realistic and large-scale applications, the classification process must be as fast and reliable as possible. Thus, each AE preprocessed waveform, $\hat{s}_q(t)$, is turned into a compact representation through a set of $\Gamma_q$ features, in both time and frequency domains. Figure 4 (a) shows an example of time domain traces of three different reference signals from tensile (red line), shear (blue line) and mixed-mode events (green line), and their Fourier spectra (Figure 4 (b)), respectively, that we have used to extract the instantaneous frequency, spectral entropy and spectral kurtosis as summarized in Table 1, we refer to as Event Descriptors (EDs).

**Table 1.** Summary of the statistical properties used in this study. We refer to these functions as Event Descriptors (EDs).

| Function | Acronym | Domain | Expression | Description | Ref. |
|---|---|---|---|---|---|
| Instantaneous Frequency | IF | Time | $IF(t) = \frac{1}{2\pi}\frac{d\phi}{dt}$ | Derivative of the phase of the analytic signal of the input | [62,63] |
| Spectral Entropy | SE | Freq. | $SE = -\frac{\sum_{m=1}^{N} P(m)\log_2(P(m))}{\log_2 N}$ | Measure of the spectral power distribution | [21,64] |
| Spectral Kurtosis | SK | Freq. | $SK \equiv \kappa(f) = \ldots \frac{m_4\{\lvert z(t)\rvert\}}{\left(m_2\{\lvert z(t)\rvert\}\right)^2} - 2$ | Describes the resemblance or difference of the shape of the spectral distribution of a signal if compared to the shape of a Gaussian bell curve | [22,65] |

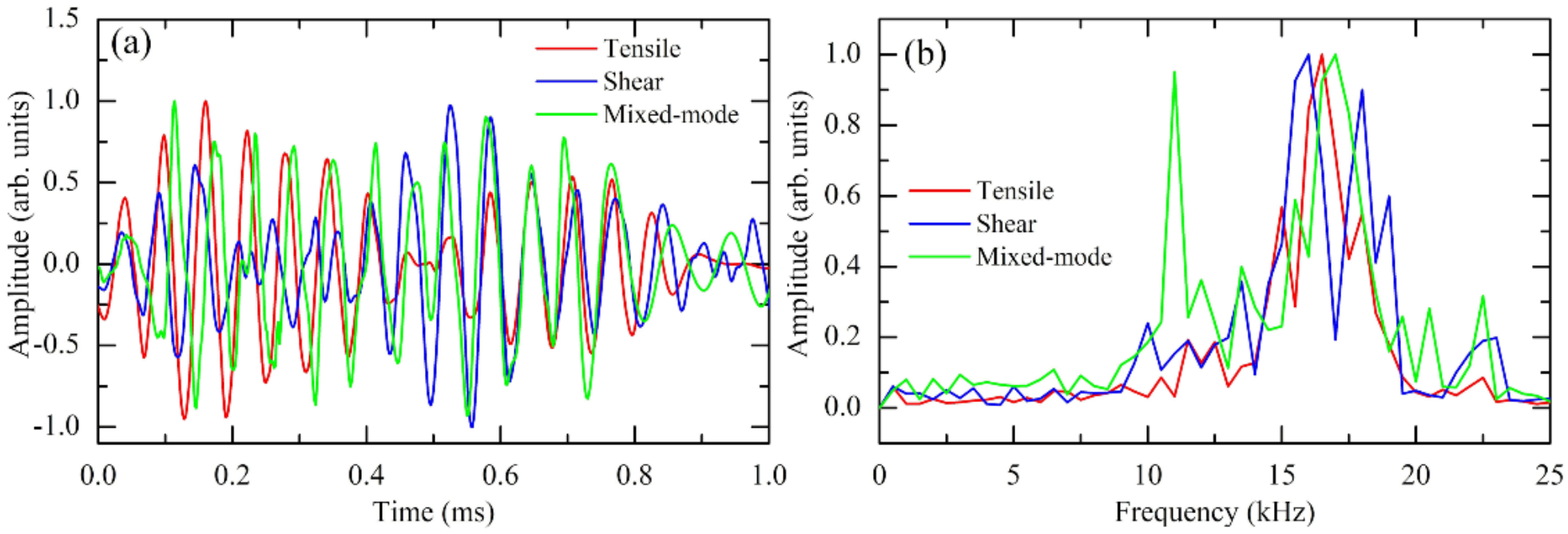

**Figure 4.** (a) Time domain representation of three investigated signals representing tensile (red), shear (blue) and mixed-mode (green), respectively. (b) Frequency spectrum representation of the same signals.

3.2.1. Instantaneous Frequency

The IF of a nonstationary signal is a time-varying parameter that relates to the average frequencies of the signal [66,67] and it is computed as the derivative of the phase of the analytic signal of the input [48]. In order to extract it, we perform the following calculations:

- Compute first the analytic signal $z(t)$ of the input, $x(t)$, such that $z(t) = x(t) + jH\{x(t)\} = A(t)e^{j\phi(t)}$, where $H(t)$ is the Hilbert Transform of $x(t)$, $A(t)$ is defined as the instantaneous power, whereas $\phi(t)$ is the instantaneous phase;
- Estimate the instantaneous frequency from the following time derivative:

$$IF(t) = \frac{1}{2\pi}\frac{d\phi}{dt} \tag{3}$$

A plot of IF as calculated of the signals of Fig. 4 (a) is shown is Fig. 5 (a). Those data extend the information of the Fourier spectra showing the range of time the modes are excited.

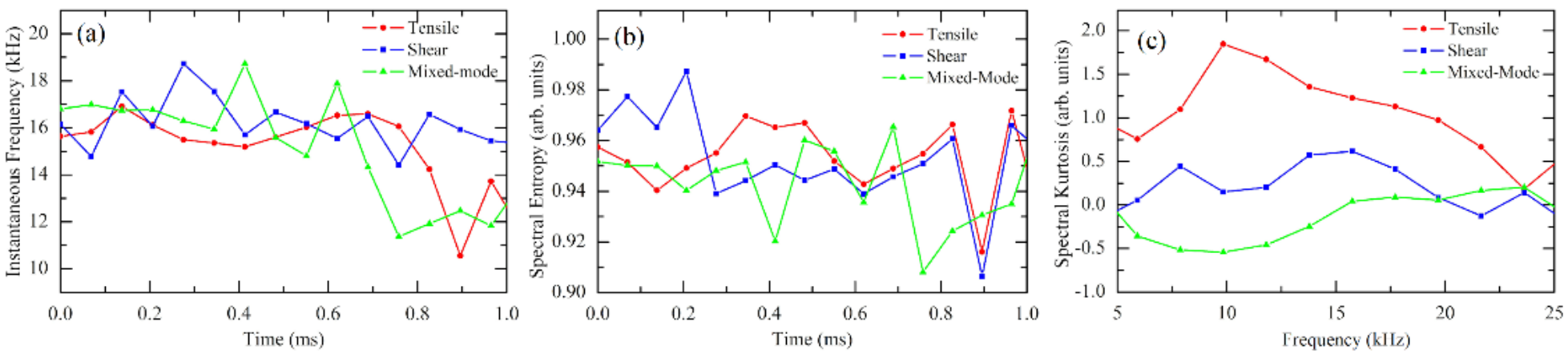

**Figure 5.** Instantaneous Frequency (a) for tensile (red line), shear (blue line), mixed-mode (green line) of the signals as represented in Figure 4 (a). Plot of Spectral Entropy (b) and Spectral Kurtosis (c), respectively.

3.2.2. Spectral Entropy

The SE [21] of a signal is a measure of its spectral power distribution. The SE treats the signal normalized power distribution in the frequency domain as a probability distribution, and calculates the Shannon entropy of it. The Shannon entropy, in the AE context, is the spectral entropy of the signal $x$. This property has been already demonstrated to be useful for features extraction in fault detection and diagnosis [68]. The equations for SE arise from the equations for the power spectrum and probability distribution for a signal. For a discrete time-varying signal $x(n)$, the power spectrum is $S(f) = |X(f)|^2$, where $X(f)$ is the discrete Fourier transform of $x(n)$. The probability distribution $P(f)$ is then:

$$P(f) = \frac{S(f)}{\sum_i S(i)} \tag{4}$$

The normalized spectral entropy $SE$ is a function of frequency and follows as:

$$SE = -\frac{\sum_{m=1}^{N} P(m)\log_2(P(m))}{\log_2 N} \tag{5}$$

where $N$ is the total number of frequency points. The denominator $\log_2 N$ represents the maximal value for the spectral entropy of white noise, which is uniformly distributed in the frequency domain. A plot of SE as calculated from the different signals of Fig. 4 (a) is shown in Fig. 5 (b). As illustrated, peculiar differences can be observed from the SE of the signals. These differences will contribute to DL-based classification although they are not easily classifiable by humans. However, from the viewpoint of power distribution among different frequency components, there is a relationship between IF and SE. We have noticed a substantial agreement with the mechanics of the events which precedes the transition from a prevalent damage process to the other one. Such transitory coexistence is well represented by the corresponding decrease in SE which puts in evidence a distinctive pattern in mixed-mode deformations when compared to tensile and shear damages. [10,13,53,55,69].

#### 3.2.3. Spectral Kurtosis

The SK [22,65] is a statistical tool that can identify the non-Gaussian behavior in the frequency domain that has been successfully used for detecting and extracting signals associated with faults in rotating mechanical systems [70]. The SK, or $\kappa(f)$, of a signal $x(t)$ can be computed based on the short-time Fourier transform (STFT) of the signal, $S(t,f)$:

$$SK \equiv \kappa(f) = \frac{\left\langle \left| S(t,f) \right|^4 \right\rangle}{\left\langle \left| S(t,f) \right|^2 \right\rangle} - 2 = \frac{m_4\left\{ \left| z(t) \right| \right\}}{\left( m_2 \left\{ \left| z(t) \right| \right\} \right)^2} - 2 \tag{6}$$

where $f \neq 0$ and $\langle \ldots \rangle$ is the time-average operator, whereas $m_4$ and $m_2$ are the raw 4-th and 2-nd order moments, respectively [22]. Figure 5 (c) shows the SK as calculated from the signals of Fig. 4 (a). The SK for tensile events (red line) is positive and higher than the corresponding function for shear (blue line) and mixed-mode (green line) events. Particularly, any SK point falling within 5-22 kHz (mostly above SK = 1.0) is likely not to be stationary and Gaussian [71]. A possible explanation is that tensile modes are characterized by a large generation of (short-lived) micro-displacements associated to the opposite movements of the crack surfaces [72] whose spectral components do not exhibit a normal distribution. The occurrence of several transient events causes multiple discontinuities in the analyzed signal which are well-captured by SK whereas other operators (for instance, power spectral density are not able to preserve such information [22]). Indeed, it is reasonable to assume that high frequency components are generated by such small discontinuities, that mainly distinguish the tensile [73,74] from shear fracture, whereas relevant frequency contributions are the results of few but larger deformations, that usually develop during the final collapse [16]. It is well-known how variations in the shape waveform during the loading process identify a change in the dominant damage mechanism of the specimen. Shear events (blue line) and mixed-mode events (green line) are characterized by smaller values of SK if compared with tensile deformations. This is because a shear event generates a relatively smaller number of transients, which corresponds to a lower value of SK, (here is slightly above 0). Interestingly, we observe that mixed-modes exhibit predominantly the lowest SK (between 5 to 20 kHz) and in particular around 11 kHz the SK = -0.5 which represents a locally stationary behavior. This is because the signal resembles a constant amplitude at the corresponding frequency (as shown in Fig. 5 (c)).

### 3.3. *Deep Learning and Bidirectional Long-Short Time Memory*

The DL represents a relatively recent branch of machine learning research [25], which attempts to model hierarchical representations behind data and classify or predict patterns by combining multiple features [10–13]. A DL network is composed of several layers, each representing a function able to apply a transformation from the input to the output. Such layers are stacked together according to the main scope (i.e. classification, prediction, recognition) and are characterized by independent properties and parameters, that affect the way they contribute to the final objective. In addition, other variables of the DL structure (i.e. number of layers, input size and characteristics, amount of elements used for the training and validation purposes) are able to impact on its performance and we commonly refer to them as model-parameters. Here, we propose a DL-based model with the aim to identify the AE among different classes of crack events. We use the Bi-LSTM network among RNNs, which is a promising solution for the problem of time series identification [75]. In a standard LSTM cell [29], the response (output) to a given input $x_t$, at time $t$, is determined by the interplay of various elements, called gates, which have different behaviors. Among them, the "forget gate" $f_t$ controls which information should be forgotten from the previous cell hidden state ($h_{t-1}$). The output gate $y_t$ highlights which information should be going to the next hidden state $h_t$.

The flow diagram is shown in Fig. 6 where an unfolded structure for a sequence of three consecutive steps ($t$ - 1, $t$, $t$ + 1) is provided.

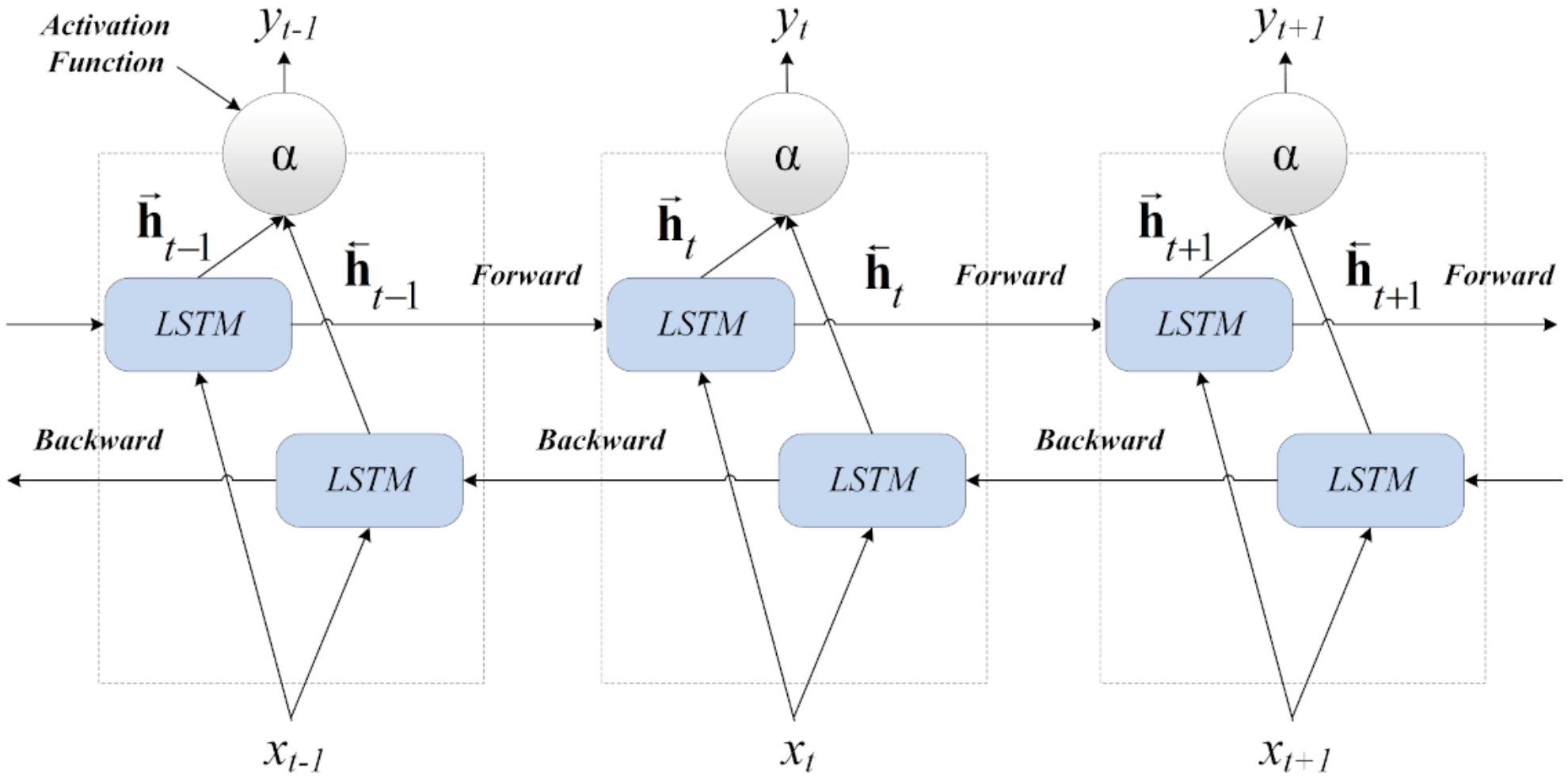

**Figure 6.** Unfolded Bi-LSTM architecture with 3 consecutive steps.

At each time step $t$, the hidden state $h_t$ is updated by a combination of: (i) current input data at the same time step $x_t$, (ii) the hidden state at the previous time-step $h_{t-1}$, (iii) the input gate $i_t$, (iv) the forget gate $f_t$, (v) the output gate $o_t$, (vi) the weight matrix W, (vii) and a memory cell ct. If we use the symbol $\longrightarrow$ to express the direction of the process, we can write:

$$\begin{aligned}
\vec{i}_t &= \alpha\left(\vec{\mathbf{W}}_i\vec{\mathbf{x}}_t + \vec{\mathbf{V}}_i\vec{\mathbf{h}}_{t-1} + \vec{\mathbf{b}}_i\right) \\
\vec{f}_t &= \alpha\left(\vec{\mathbf{W}}_f\vec{\mathbf{x}}_t + \vec{\mathbf{V}}_f\vec{\mathbf{h}}_{t-1} + \vec{\mathbf{b}}_f\right) \\
\vec{o}_t &= \alpha\left(\vec{\mathbf{W}}_o\vec{\mathbf{x}}_t + \vec{\mathbf{V}}_o\vec{\mathbf{h}}_{t-1} + \vec{\mathbf{b}}_o\right) \\
\vec{c}_t &= \vec{f}_t \odot \vec{c}_{t-1} + \vec{i}_t \odot \tanh\left(\vec{\mathbf{W}}_c\vec{\mathbf{x}}_t + \vec{\mathbf{V}}_c\vec{\mathbf{h}}_{t-1} + \vec{\mathbf{b}}_c\right) \\
\vec{h}_t &= \vec{o}_t \odot \tanh\left(\vec{c}_t\right)
\end{aligned} \tag{7}$$

Where $\alpha$ is the activation function, and $\odot$ represents the element-wise product. The Bi-LSTM is able to process the input sequence in both directions (forward and backward) with two separate hidden layers in order to account for the full input context. The following equations define the corresponding hidden layer function, whereas the $\longrightarrow$ and $\longleftarrow$ denote the forward and backward process, respectively. For the forward process we consider Eq. (20), whereas for the backward process we have:

$$\begin{aligned}
\overleftarrow{i}_t &= \alpha\left(\overleftarrow{\mathbf{W}}_i\overleftarrow{\mathbf{x}}_t + \overleftarrow{\mathbf{V}}_i\overleftarrow{\mathbf{h}}_{t+1} + \overleftarrow{\mathbf{b}}_i\right) \\
\overleftarrow{f}_t &= \alpha\left(\overleftarrow{\mathbf{W}}_f\overleftarrow{\mathbf{x}}_t + \overleftarrow{\mathbf{V}}_f\overleftarrow{\mathbf{h}}_{t+1} + \overleftarrow{\mathbf{b}}_f\right) \\
\overleftarrow{o}_t &= \alpha\left(\overleftarrow{\mathbf{W}}_o\overleftarrow{\mathbf{x}}_t + \overleftarrow{\mathbf{V}}_o\overleftarrow{\mathbf{h}}_{t+1} + \overleftarrow{\mathbf{b}}_o\right) \\
\overleftarrow{c}_t &= \overleftarrow{f}_t \odot \overleftarrow{c}_{t+1} + \overleftarrow{i}_t \odot \tanh\left(\overleftarrow{\mathbf{W}}_c\overleftarrow{\mathbf{x}}_t + \overleftarrow{\mathbf{V}}_c\overleftarrow{\mathbf{h}}_{t+1} + \overleftarrow{\mathbf{b}}_c\right) \\
\overleftarrow{h}_t &= \overleftarrow{o}_t \odot \tanh\left(\overleftarrow{c}_t\right)
\end{aligned} \tag{8}$$

Then, the complete Bi-LSTM hidden element representation $h_t$ is the concatenated vector of the outputs of forward and backward processes, such that $h_t = \vec{h}_t \oplus \overleftarrow{h}_t$.

Here, the two sub-layers compute forward $\vec{h}$ and backward $\overleftarrow{h}$ hidden sequences, respectively, which are then combined to compute the output sequence y as the concatenation of both input sequences, $\mathbf{y}_t = h_t \mathbf{x}_t$ . Bi-LSTM units are a key element in our DL model because of their ability to classify, process and predict time series with time lags of unknown duration. Relative insensitivity to gap length gives an advantage to Bi-LSTMs over alternative RNNs, hidden Markov models and other sequence learning methods [76].

### *3.4. DL-Based Event-Type Discrimination*

A number of concrete (e.g. fibre reinforced concrete, pre-stressed concrete) are used in literature for analyzing failure mechanisms using AE events. An acoustic wave comprises of different sub-events, which can be categorized into few main classes, where the waveform shape is directly connected to the type of deformation, as also explained in Section 3.1.

To determine the crack type, various statistical properties of the signal have been considered (see Section 3.2). Thus, we use a combination of these properties to make a DL network able to distinguish among such different classes of events.

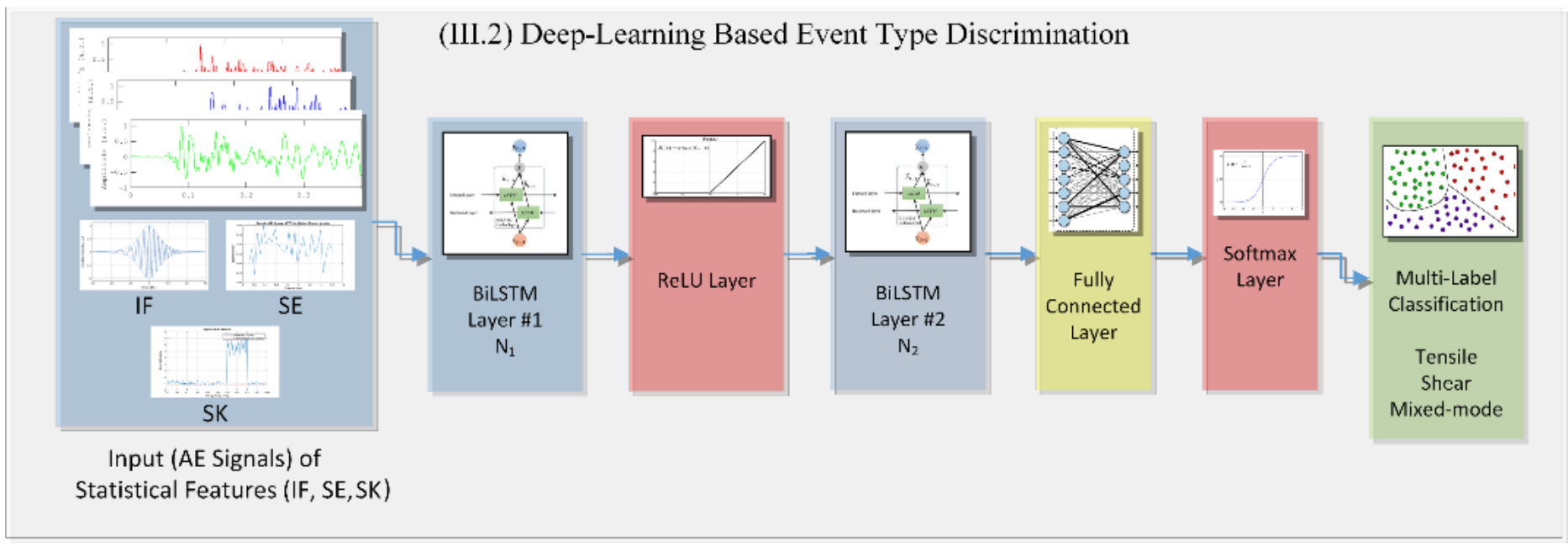

**Figure 7.** Proposed network for DL-based event type discrimination.

The internal structure of block (III.2) in Fig. 7 is organized by means of the interposition of functional layers with the aim to provide convergence of training and regularization methodology to ensure that a flexible deep learning model finds solutions with good predictive performance. Generalization to data outside the training set is the key objective of predictive machine learning methods. In this block, the features input array, Γ, from AE signals is used to train the DL network and evaluate multiple crack events. We use Bi-LSTM units as a key ingredient for the Multi-Label classification. In our scenario, events are categorized into [13]:

- Tensile event [10,77,78];
- Shear event [13,79–81];
- Mixed mode [53,56,57,82–85].

Once the input training data Γ are collected at the bottom of the network (first block on the left), they are used to feed a stack composed of several layers. To start with, we have a cascade of different Bi-LSTM layers composed of $N_1$ and $N_2$ hidden units, respectively. The output of the first Bi-LSTM layer is then used as input for the subsequent activation function layer which is represented by a Rectified Linear Unit (ReLU) [86]. ReLU has been demonstrated to be particularly valuable for the study of the classification or prediction once applied on time sequences, and has been shown to outperform more conventional approaches based on previous sigmoid and tanh functions [87].

In our DL model, each element of the output of ReLU layer flows through a second Bi-LSTM block having $N_2$ cells, then a fully-connected layer is interposed. It multiplies the input by a weight matrix and then adds a bias vector. The default for the initial weights is a Gaussian distribution with mean zero and unit variance. The default for the initial bias is 0. Given the nature of the classification task, a subsequent Softmax layer [88] constitutes an essential part at the top of the DL network. Here, the Softmax function $y_r$ is applied to the input $x$ having the following expression:

$$y_r(x) = \frac{\exp\left(a_r(x)\right)}{\sum_{j=1}^{k} \exp\left(a_j(x)\right)}, \text{ where } 0 \le y_r \le 1 \text{ and } \sum_{j=1}^{k} y_j = 1 \tag{10}$$

where $a_r(x)$ represents the conditional probability of the sample given class $r$ (we have $r = 1, 2, 3$). If we set $y_r = P(c_r \mid x, \theta)$, and $a_r(x) = \ln\left(P(x, \theta \mid c_r) P(c_r)\right)$, with $P(c_r)$ being the class prior probability [89], we obtain:

$$P(c_r \mid x, \theta) = \frac{P(x, \theta \mid c_r) P(c_r)}{\sum_{j=1}^{k} P(x, \theta \mid c_j) P(c_j)} \tag{11}$$

Specifically, in multinomial logistic regression and linear discriminant analysis, the input to the function is the result of $k$ distinct linear functions, and the predicted probability for the class $r$ given a sample vector $x$ and a weighting vector θ, such that $0 \le P(c_r \mid x, \theta) \le 1$ and $\sum_{j=1}^{k} P(c_j \mid x, \theta) = 1$. Finally, a classification layer [88] is the last (top of the DL network) and it computes the cross entropy (CE) loss [87] for multiclass classification problems with mutually exclusive classes as [88]:

$$CE = \sum_{i=1}^{Q} \sum_{r=1}^{R=3} t_{ir} y_{ir} \tag{12}$$

where $Q$ is the number of samples, $r$ is the number of classes, $t_{ir}$ is the indicator function that establishes if the $i$-th sample belongs to the $r$-th class, and $y_{ir}$ is the output of the $i$-th sample for class $r$, which, in this case, is the value from the Softmax function. Basically, this is the probability that the network associates the $i$-th input with class $r$. The output of the top layer is the most probable class according to the way the DL-network has been trained.

3.4.1. Training model

Although the proposed deep stacked Bi-LSTM network is composed of several layers $L$, it can be still abstracted as a function $F$ of input $x$ with parameters matrix Θ:

$$[\hat{y}] = F(x, \Theta) \tag{13}$$

where $\hat{y}$ is the estimated output, and Θ represents all the weights $w_i$ and biases $\gamma_i$ for all the layers $(i = 1, \ldots L)$:

$$\Theta = [w_i, \gamma_i]_{i=1,\ldots L} \tag{14}$$

By using the mini-batch Stochastic Gradient Descent method (SGDM) [87], Θ can be optimized to minimize the CE losses. By applying the training algorithm Back-Propagation Through Time (BPTT) [14], the following numerical computation is iteratively performed to update the parameters to lower the loss function toward the optimum:

$$\bar{\Theta} = \Theta - r_l \frac{\partial CE(\Theta)}{\partial \Theta} \tag{15}$$

where we use gradient descent to adjust the learning rate $r_l$ automatically. Early Stopping is also applied to prevent overfitting [16]. Data samples are prepared in advance for the training process. The construction and the training process of the model are both implemented using the latest TensorFlow [90]. Totally 15,000 digitized samples are gathered for the training of the BiLSTM model and 1,650 are used for testing purposes.

3.4.2. Model-parameter settings

There are multiple model-parameters in our proposed method that have an effect on the performance: (i) the size of the training dataset $Q$, (ii) the combination $\lambda_r$ of available inputs $\Gamma_q$, (iii) the total number $d_{LSTM} = N_1 + N_2$ of memory cells in the model, which affects the complexity of the network, and (iv) the number of epochs $\mathcal{E}$ used to train the network. As stated above, being that the size of $Q$ is considered large enough for the training, we have to find the optimal combination of $\lambda_r$, $d_{LSTM}$ and $\mathcal{E}$ values for the most accurate classification of crack events.

| Available inputs | Signal | IF | SE | SK | Size of input dataset $\Gamma_q$ |
| --- | --- | --- | --- | --- | --- |
| $\lambda_1$ | yes | no | no | no | $Q \times 1 \times N_{ED}$ |
| $\lambda_2$ | no | yes | no | no | $Q \times 1 \times N_{ED}$ |
| $\lambda_3$ | no | yes | yes | no | $Q \times 2 \times N_{ED}$ |
| $\lambda_4$ | no | yes | no | yes | $Q \times 2 \times N_{ED}$ |
| $\lambda_5$ | no | yes | yes | yes | $Q \times 3 \times N_{ED}$ |

**Table 2.** Description of the components of each combination of inputs $\lambda_1$ - $\lambda_5$ used to train the DL-network.

## 4. Results and Discussions

An extensive preliminary study has been carried out to identify, collect and categorize tensile, shear and mixed-mode as the recognized classes of deformation. The results we have achieved in processing the AE data are qualitatively similar for all the specimens under test. Thus, we discuss in detail the results from one dataset which represents an ensemble of crack events from multiple specimens. Only a minor part of such measurements has been improved via data augmentation [91,92] to ensure the required balance between the different types of failure. For a given event $\tilde{s}_q(t)$ in any class (i.e. tensile, shear and mixed-mode), an array of $\Gamma_q$ EDs is calculated and used to populate the training dataset. Each ED is a series composed of $N_{ED}$ elements. Such input dataset is therefore constituted of $Q$ x number-of-EDs x $N_{ED}$, whose details are provided in Table 2. Standardization is applied on data to ensure zero mean and unit variance and to ease the training process. In order to avoid over-fitting during the learning process and improve classification accuracy, the dataset was divided into training and validation quotas at an 80/20 ratio.

### *4.1. Performance optimization of Model-parameters*

We use a training dataset having $Q$ = 15,000, mini-batch sample size $m$ = 1500, initial learning rate $r_l = 0.001$ using SGDM optimizer. We evaluate a different number of hidden units of the network $d_{LSTM}$ = 200, 500 and 800, for various configurations of input data $\lambda_1$ - $\lambda_5$ (see Table 2). For all the applications, we consider a maximum number of epochs $\mathcal{E}$ = 220, to but to avoid overfitting the learning process stops when the cross-entropy validation loss (i.e. CE in Eq. (12)) does not decrease after 30 epochs [87]. We conducted several numerical tests to determine the optimal configuration.

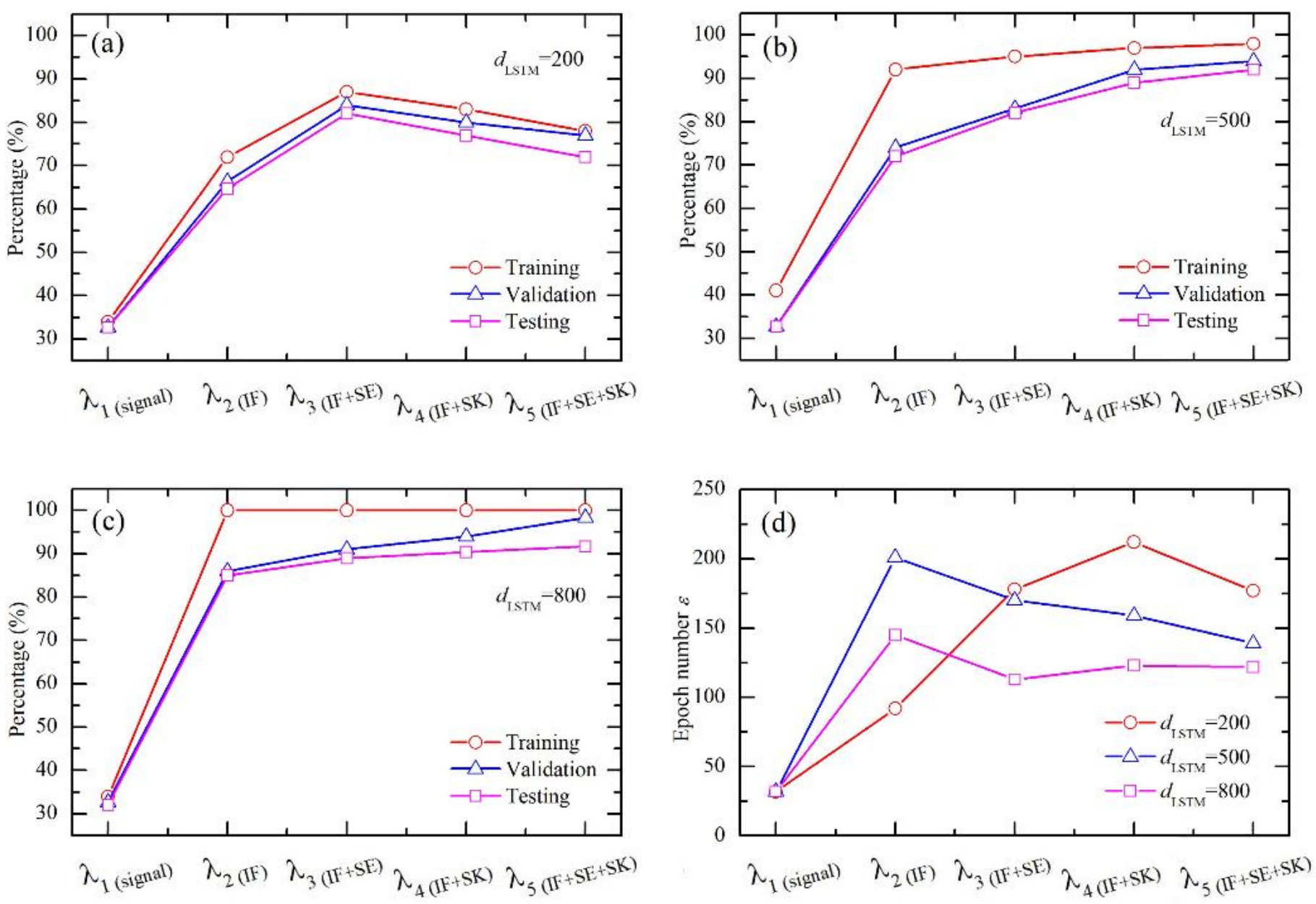

**Figure 8.** (a)-(d) Effects of model-parameters in tuning of Bi-LSTM network with a dataset Q = 10.000 AE events. (a) Performance of training, validation and testing with different input configurations ($\lambda_1$-$\lambda_5$) using $d_{LSTM}$ = 200 (a), $d_{LSTM}$ = 500 (b), $d_{LSTM}$ = 800 (c), respectively. (d) Trend of epoch number $\mathcal{E}$ = 300 as a function of the different input configurations ($\lambda 1$-$\lambda 5$) and for $d_{LSTM}$ = 200, 500 and 800, respectively.

Figure 8 (a-c) summarize the effect of the number of $d_{LSTM}$ (or hidden units) on classification accuracy (i.e. ratio between the number of correctly classified events and the total number of events in the considered dataset, in percentage) for the training, validation and test dataset (testing) of the proposed model ($d_{LSTM}$ = 200 in Fig. 8 (a), $d_{LSTM}$ = 500 in Fig. 8 (b), and $d_{LSTM}$ = 800 in Fig. 8 (c)). Here, we can observe that the performance of this model is better as the number $d_{LSTM}$ increases under the defined range ($d_{LSTM}$ = [200-800]), while it does not undergo saturation or degradation as the scale becomes larger. A value of $d_{LSTM}$ = 500 provides the best testing performance with an accuracy of 92% (by combining IF, SE and SK). The reason is that the Bi-LSTM model with fewer LSTM cells suffers from lack of capability of long-term memory, which will result in under-fitting. Whereas, a model too large (i.e. higher number of hidden cells) will always lead to increasing difficulty due to over-fitting. In Fig. 8 (d), we show the validation performance of Bi-LSTM models in terms of epochs (required to complete the training) for different numbers of cells and inputs ($\lambda 1$-$\lambda 5$). After the training is completed, the model has learned the patterns necessary for the classification of the input events. As expected, the epoch number $\mathcal{E}$ of each model is different, depending on the choice of $d_{LSTM}$, and the input configuration $\lambda_r$. We also observed that the operation time of each epoch with different $\lambda_r$ is similar. This is because the computational complexity is mainly affected by both the size $Q$ of the training dataset, and the $d_{LSTM}$, the complexity of the network, which mostly impact on the process of weight updating and gradient propagation. We only adopt the $\Theta$ that makes the lowest $J(\Theta)$ on the validation dataset for each model and we collect the value of loss function for both training $J_{TRN}(\Theta)$ and validation $J_{VAL}(\Theta)$. The training, validation and testing accuracy increase with $\lambda_r$ when $d_{LSTM}$ is higher than 200 LSTM cells. For $d_{LSTM}$ = 200, a local minimum in validation loss is observed with $\lambda_r$ = $\lambda_3$ (which implies using a combination of IF and SE for each acoustic signal). Epoch number $\mathcal{E}$ tends to decrease with $d_{LSTM}$. As expected, the $\mathcal{E}$ required for the calculation of the four different statistical measures $\lambda_5$ (which implies using IF, SE and SK for each signal) is the lowest. Figure 9 shows the learning error by the value of the loss function for both training (Fig. 9(a)) and validation (Fig. 9(b)) tasks.

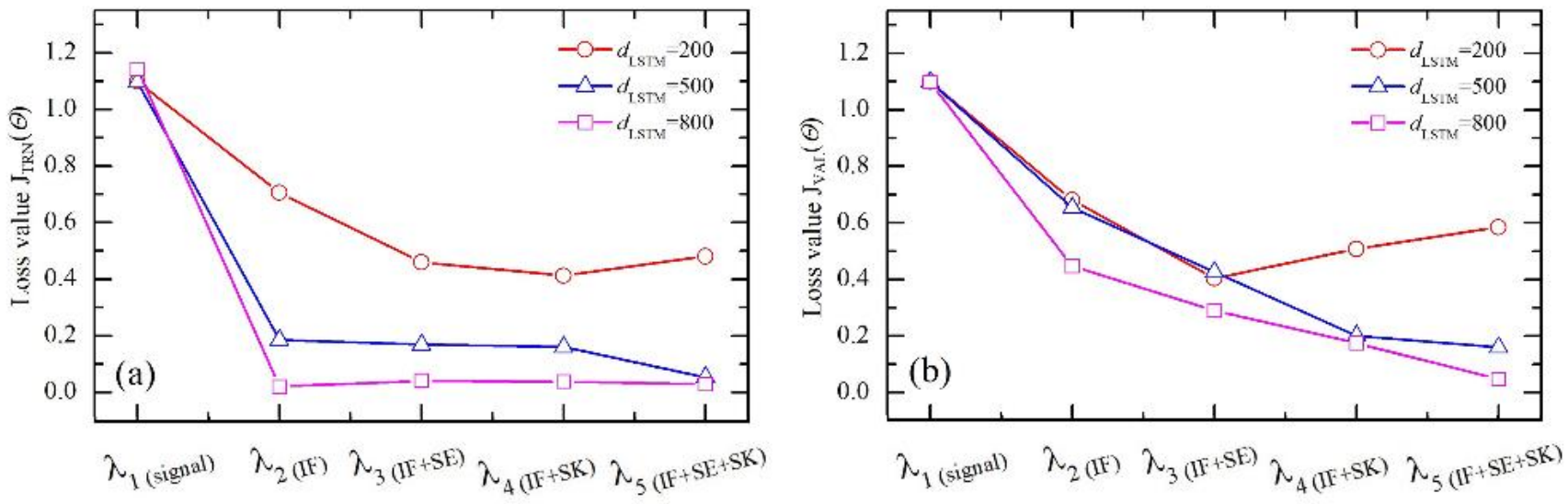

**Figure 9.** (a) Learning error calculated by the value of loss function for training $J_{TRN}(\Theta)$ and validation (b) $J_{VAL}(\Theta)$ tasks, respectively, under different input combinations $\lambda_1$ - $\lambda_5$, and for different values of $d_{LSTM}$.

The results for the different combinations of inputs $\lambda_1$ - $\lambda_5$ and a higher classification accuracy are preferred over learning time. Based on these analyses, we set the number of the LSTM cells in our Bi-LSTM model to $d_{LSTM}$ = 500, since it provides the lowest validation loss ($J_{VAL}(\Theta)$), and choose the input configuration $\lambda_5$ where we take into account IF, SE and SK as representative properties of the signal.

## 5. Conclusions

In summary, this work describes the strategy to develop deep neural network with Bi-LSTM for the classification of crack originating time domain AE. The key ingredient is a proper set of EDs we found to be the instantaneous frequency, spectral entropy, and spectral kurtosis. In particular, our results show that the use of EDs as input of a DL-based network gives rise to an accuracy of 92%. We wish to highlight that this approach implemented for fault diagnosis/classification problem of concrete can be generalized for other materials or for predicting failures with minimum adjustments and moderate re-training [87].

In addition, the methods presented here provide capabilities for real time monitoring in large scale applications and might be suitable to be implemented in the sensing layer of IoT-based monitoring systems [93] and for digital twin.

**Author Contributions:** G.S., and F.G. developed the idea; G.S. implemented the algorithm and the methodology supported by A.L.C. and F.G.; A.L.C., M.Ch., and G.F. coordinated the work; F.L.M. and C.S. performed the experimental measurements. G.S., F.G. and G.F. wrote the paper with input from R.T. and A.L.C. In addition, M.Ch., R.T. and M.Ca. reviewed the paper, formalized theoretical concepts and provided figures and reference check. The evaluation part has been accomplished by G.S., F.G. and G.F. All authors have read and contributed to the last version of the manuscript.

**Funding:** The authors acknowledge the financial support from Petaspin association (www.petaspin.com). A.L.C., M.Ch., and G.F. thank the project "Pipeline for Advanced Contrast Enhancement (PACE) for Enhancement effectiveness of chest X-ray for monitoring COVID-19 patients (EX-COVID)", grant n. J45F21000460001, funded from the Ministry of Education, Universities and Research (MIUR), Italy.

**Acknowledgments:** This work was partially supported by PETASPIN association and MARIS s.c.a.r.l. The authors thank the NGT test laboratory and the engineers Gennaro Angotti, Cinzia Angotti and Dino Padula for the tests carried out in their structure.

## Appendix A

The training and testing dataset are available online (https://github.com/giuliosiracusano/deeplearning_crack_classification).